\documentclass{article}

\usepackage[preprint]{neurips_2026}

\usepackage[utf8]{inputenc}
\usepackage[T1]{fontenc}
\usepackage{hyperref}
\usepackage{url}
\usepackage{booktabs}
\usepackage{amsfonts}
\usepackage{amsmath}
\usepackage{amssymb}
\usepackage{mathtools}
\usepackage{amsthm}
\usepackage{nicefrac}
\usepackage{microtype}
\usepackage{xcolor}
\usepackage{graphicx}
\usepackage{algorithm}
\usepackage{algorithmic}
\usepackage{tabularx}
\usepackage[capitalize,noabbrev]{cleveref}
\usepackage[textsize=tiny]{todonotes}

\theoremstyle{plain}

\theoremstyle{definition}

\theoremstyle{remark}

\newcommand{\vfield}{v_\theta}
\newcommand{\R}{\mathbb{R}}
\newcommand{\E}{\mathbb{E}}
\newcommand{\loss}{\mathcal{L}}

\title{FLUX: Geometry-Aware Longitudinal Flow Matching\\with Mixture of Experts}

\author{
\normalfont Josue Ortega Caro$^{1,*}$ \quad
\normalfont Yongxu Zhang$^{1}$ \quad
\normalfont Hanna M. Batchelor$^{1}$ \\
\normalfont Sizhuang He$^{1}$ \quad
\normalfont Jessica Cardin$^{1}$ \quad
\normalfont Shreya Saxena$^{1,*}$ \\
$^{1}$Yale University \\
$^{*}$Correspondence: \texttt{josue.ortegacaro@yale.edu}, \texttt{shreya.saxena@yale.edu}
}
\begin{document}

\maketitle

\begin{abstract}
Many biological systems evolve through continuous local dynamics while switching between latent regimes defined by learning, stimulus context, internal state, or developmental stage. These processes are often observed only as unpaired longitudinal snapshots: the same cells, neurons, or animals are not tracked as matched trajectories, even though population states are sampled across successive stages. This creates two coupled challenges. First, trajectories must respect curved low-dimensional manifolds embedded in high-dimensional biological measurements. Second, the model must identify when the transport mechanism itself changes. We introduce \textbf{FLUX} (\textbf{Fl}ow matching for \textbf{U}npaired longitudinal data with mi\textbf{X}ture-of-experts), a geometry-aware longitudinal flow-matching framework for joint transport modeling and unsupervised regime discovery. FLUX learns a data-dependent metric from pooled labeled and unlabeled observations, uses that metric to construct geometry-aware conditional paths between adjacent marginals, and decomposes the resulting velocity field into sparse expert vector fields selected by a Straight-Through Gumbel-Softmax router. Across manifold controls, a regime-switching Lorenz system, widefield cortical calcium imaging during associative learning, and embryoid body single-cell differentiation, FLUX reconstructs longitudinal transport while recovering interpretable regime structure. In neural data, the router separates early/intermediate training from late training, coinciding with the behavioral divergence of CS$+$ and CS$-$ lick indices. In embryoid body differentiation, FLUX separates expression-evolution regimes associated with pluripotent and more differentiated cell populations. Ablations show that mixture-of-experts routing alone is insufficient: FLUX without geometric learning can fit local transport but fails or weakens regime discovery when regimes are encoded in local dynamics. These results suggest that geometry-aware velocity decomposition provides a general strategy for discovering latent biological state transitions frr32om unpaired longitudinal snapshots.

\end{abstract}

\section{Introduction}
Many biological systems are observed only as unpaired population snapshots. In widefield calcium imaging across days of learning, animals are imaged repeatedly but neurons drift, sessions vary, and trial structure is not aligned across animals. In single-cell RNA sequencing, measurement destroys the cell, so the same cells cannot be tracked across timepoints. In these cases, the data are sequences of marginal distributions $\mu_0,\mu_1,\ldots,\mu_{T-1}$, rather than tracked trajectories, yet the underlying biological process is continuous: cells differentiate through intermediate states, and neural population activity evolves smoothly within and across learning phases. The inferential goal is therefore recovering a dynamical process that is consistent with all observed marginals. 

For biological systems, longitudinal transport is only part of the scientific problem. Neural circuits and differentiating cell populations often change the rules of their dynamics over time. During associative learning, early activity may reflect broad stimulus--reward expectation, whereas late activity may reflect stimulus-specific discrimination. During development, transcriptional dynamics may transition from pluripotent programs to lineage-committed or differentiated programs. These transitions are regimes of the dynamics: they are changes in the local transport mechanisms or \emph{flows}, not necessarily changes that are visible as clean spatial clusters in the raw observation space. A useful model should therefore infer both the continuous population transport and the discrete latent regime structure organizing that transport.

A second challenge is geometric. High-dimensional biological measurements often concentrate near low-dimensional nonlinear manifolds due to structural or biophysical constraints underlying the biological process. Simple approximations such as Euclidean interpolants between unpaired endpoints can leave the data manifold, pass through low-density regions, and provide velocity targets that do not reflect the local geometry of the population, potentially violating biophysical constraints. This issue is especially acute in longitudinal snapshot settings because trajectory labels are sparse: only a subset of samples may be associated with timepoint labels or experimental stages, here specific marginals, while many additional observations may still carry information about the manifold. Manifold learning can exploit these unlabeled observations to estimate geometry, and this geometry can then constrain trajectory inference using the labeled longitudinal marginals. 

Flow matching provides a simulation-free objective for learning continuous-time velocity fields from prescribed conditional paths \citep{lipman2023flow}. Conditional flow matching and minibatch optimal-transport couplings improve two-marginal transport \citep{tong2024improving,pooladian2023multisample}, while longitudinal extensions chain transports across multiple timepoints \citep{islam2025longitudinal,albergo2023multimarginal}. Riemannian and metric flow-matching approaches replace Euclidean paths with geometry-aware paths when the data lie on curved manifolds \citep{chen2024riemannian,kapusniak2024metric}. These advances address transport, but they do not directly turn the learned velocity field into a regime-discovery model for biological systems.

\begin{figure}[t]
    \centering
    \includegraphics[width=\textwidth]{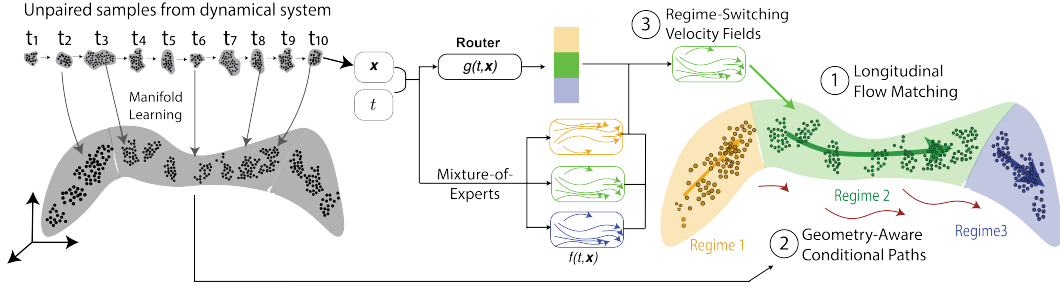}
    \caption{\textbf{Overview of FLUX.} Longitudinal data are observed as unpaired population snapshots. FLUX builds on Metric Flow Matching by using a learned data-dependent geometry to construct manifold-aware conditional paths, and extends this construction to ordered multi-marginal sequences with a regime-switching velocity field. A router $g$ computes expert logits, $M$ expert velocity networks predict candidate vector fields in parallel, and Straight-Through Gumbel-Softmax selects the active expert. The resulting expert assignment provides an unsupervised candidate regime label.}
    \label{fig:overview}
    \vspace{-0.2cm}
\end{figure}

We introduce FLUX (Flow matching for Unpaired longitudinal data with miXture-of-experts), a geometry-aware longitudinal flow-matching model with sparse velocity-field decomposition (Figure \ref{fig:overview}). FLUX first learns a data-dependent metric from pooled observations, including unlabeled samples when available. It then uses a bend network to construct geometry-aware conditional paths between adjacent marginals. Finally, it trains a mixture-of-experts velocity field in which a router $g$ selects among $K$ expert vector fields. The resulting expert assignment is interpreted as an unsupervised candidate regime label because it decomposes the transport mechanism itself rather than clustering static observations after the fact.

Our contributions are:
\begin{itemize}
    \item We extend geometry-aware flow matching to longitudinal marginal chains and emphasize the role of unsupervised manifold learning when trajectory labels are sparse, but unlabeled observations are available.
    \item We introduce a regime-switching velocity field in which sparse expert routing discovers latent dynamical regimes directly from local transport structure.
    \item We show that geometry is very helpful towards regime discovery, especially when regimes are switching due to changes in local dynamics.
    \item We demonstrate biological relevance in two diverse settings: FLUX separates early/intermediate from late neural training periods during the emergence of CS$+$/CS$-$ behavioral discrimination, and it recovers expression-evolution differences between pluripotent and more differentiated embryoid-body cell populations.
\end{itemize}

\section{Related Work}
\paragraph{Flow matching and longitudinal transport.}
Neural ODEs parameterize continuous-time dynamics but require differentiating through ODE solves during training \citep{chen2018neural}. Flow matching avoids this cost by regressing a velocity field onto conditional probability paths \citep{lipman2023flow}. Optimal-transport and minibatch couplings straighten these flows and improve generative performance \citep{tong2024improving,pooladian2023multisample}. Related interpolation-based generative frameworks include rectified flow and stochastic interpolants, which also learn transport dynamics from prescribed probability paths \citep{liu2023flow,albergo2023stochastic}. Longitudinal settings require transport across more than two marginals: Waddington-OT reconstructs developmental trajectories using static optimal transport \citep{schiebinger2019optimal}, TrajectoryNet learns continuous normalizing flows from snapshot single-cell data \citep{tong2023trajectorynet}, and recent longitudinal flow-matching and stochastic-interpolant methods chain or couple multiple marginals for trajectory modeling \citep{islam2025longitudinal,albergo2023multimarginal}. FLUX adopts adjacent-marginal chaining but replaces Euclidean conditional paths with learned geometry-aware paths and adds velocity-field decomposition for regime discovery.

\paragraph{Geometry-aware biological trajectory models.}
When data concentrate near a curved manifold, Euclidean interpolants can pass through low-density regions and produce unrealistic velocity targets. Riemannian flow matching extends flow matching to known manifolds with tractable geodesics \citep{chen2024riemannian}, while Metric Flow Matching learns data-dependent metrics and approximate geodesic interpolants \citep{kapusniak2024metric}. Deep-kernel metric models provide a practical alternative in higher-dimensional observations by learning a feature map before applying a kernelized distance-to-manifold score. In biological trajectory inference, MIOFlow combines manifold learning, optimal transport, and neural ODEs to model continuous population dynamics from static snapshots \citep{huguet2022manifold}. Other generative models for time-series single-cell data include PRESCIENT and scNODE, which learn continuous or stochastic dynamics from population snapshots but do not decompose the learned velocity field into regimes \citep{yeo2021prescient,zhang2024scnode}. FLUX uses geometry-aware conditional paths inside a longitudinal marginal chain and treats the learned velocity field itself as the target for regime decomposition.

\paragraph{Mixture-of-experts and regime discovery.}
Mixture-of-experts models route inputs to specialized subnetworks, increasing functional capacity while encouraging specialization \citep{shazeer2017outrageously}. Switch Transformers popularized sparse top-1 routing and load-balancing strategies for stable expert specialization \citep{fedus2022switch}. Gumbel-Softmax enables differentiable approximations to discrete routing decisions \citep{jang2017categorical}. Classical regime-discovery models include hidden Markov models \citep{rabiner1989tutorial}, recurrent switching linear dynamical systems \citep{linderman2017rslds}, and switching recurrent neural networks \citep{zhang2024srnn}. These methods typically assume tracked trajectories or sequential observations. FLUX instead discovers regimes from unpaired population snapshots by decomposing a learned flow-matching velocity field.

\section{Methods}
\subsection{Longitudinal Flow Matching}

We observe $T$ ordered marginal distributions $\mu_0,\mu_1,\ldots,\mu_{T-1}$ at observation times $s_0<\cdots<s_{T-1}$. Each marginal contains unpaired samples
\begin{equation}
    \mathcal{D}_k=\{\mathbf{x}_k^{(i)}\}_{i=1}^{N_k},
    \qquad
    \mathbf{x}_k^{(i)}\in\R^d,
\end{equation}
where $d$ is the ambient feature dimension. Samples in consecutive marginals are not assumed to correspond to the same cell, neuron, trial, or individual. The goal is to learn a time-dependent velocity field
\begin{equation}
    \vfield:[0,1]\times\R^d\rightarrow\R^d,
    \qquad
    \frac{d\mathbf{x}(t)}{dt}=\vfield(t,\mathbf{x}(t)),
\end{equation}
whose flow map $\phi_{a\rightarrow b}$ transports the marginal sequence:
\begin{equation}
    (\phi_{0\rightarrow t_k})_{\#}\mu_0 \approx \mu_k,
    \qquad
    t_k=\frac{k}{T-1}.
\end{equation}
Here $(\cdot)_{\#}$ denotes pushforward, $t\in[0,1]$ is normalized model time, and $t_k$ is the normalized time of marginal $k$.

FLUX trains on adjacent marginal pairs. For pair $(\mu_k,\mu_{k+1})$, endpoints $(\mathbf{x}_k,\mathbf{x}_{k+1})$ are sampled from a coupling
\begin{equation}
    \pi_k \in \Pi(\mu_k,\mu_{k+1}),
\end{equation}
and a local interpolation time $\alpha\sim\mathrm{Uniform}(0,1)$ is mapped to global model time by
\begin{equation}
    t_{k,\alpha}
    =
    \frac{k+\alpha}{T-1}.
\end{equation}
Thus, two-marginal conditional paths are applied across adjacent pairs to learn one longitudinal velocity field over the full marginal chain.

\subsection{Geometry-Aware Conditional Paths}

The geometry-aware path construction in FLUX follows Metric Flow Matching~\citep{kapusniak2024metric}. Standard conditional flow matching uses the Euclidean path
\begin{equation}
    \mathbf{z}^{\mathrm{Euc}}_{k,\alpha}
    =
    (1-\alpha)\mathbf{x}_k+\alpha\mathbf{x}_{k+1},
    \qquad
    \dot{\mathbf{z}}^{\mathrm{Euc}}_{k,\alpha}
    =
    \mathbf{x}_{k+1}-\mathbf{x}_k .
\end{equation}
For data near a curved manifold, this path can pass through low-density regions. FLUX instead uses a learned geometry model $G$ to define low-energy paths. Rather than solving exact geodesics for every endpoint pair, a bend network $B$ parameterizes an approximate geometry-aware path
\begin{equation}
    \gamma_{k}(\alpha)
    =
    B(\mathbf{x}_k,\mathbf{x}_{k+1},\alpha;G),
    \qquad
    \gamma_{k}(0)=\mathbf{x}_k,
    \qquad
    \gamma_{k}(1)=\mathbf{x}_{k+1}.
\end{equation}
The flow-matching training point and target tangent are
\begin{equation}
    \mathbf{z}_{k,\alpha}=\gamma_{k}(\alpha),
    \qquad
    \dot{\mathbf{z}}_{k,\alpha}
    =
    \partial_\alpha \gamma_{k}(\alpha).
\end{equation}

The metric is therefore not an additional input to the velocity network; it changes the conditional paths, training locations, and target tangents used for supervision. FLUX uses this Metric Flow Matching geometry layer for longitudinal multi-marginal snapshot modeling via sparse set of vector fields.

\subsection{Regime-Switching Velocity Fields}

FLUX decomposes the longitudinal velocity field into $M$ expert vector fields,
\begin{equation}
    f_m:[0,1]\times\R^d\rightarrow\R^d,
    \qquad m=1,\ldots,M .
\end{equation}
A router
\begin{equation}
    g:[0,1]\times\R^d\rightarrow\R^M
\end{equation}
maps a time and state to expert logits
\begin{equation}
    g(t,\mathbf{x})
    =
    \left(\ell_1(t,\mathbf{x}),\ldots,\ell_M(t,\mathbf{x})\right).
\end{equation}
The routing weights $\mathbf{w}(t,\mathbf{x})\in\Delta^{M-1}$ define
\begin{equation}
    \vfield(t,\mathbf{x})
    =
    \sum_{m=1}^{M}
    w_m(t,\mathbf{x}) f_m(t,\mathbf{x}).
    \label{eq:flux_velocity}
\end{equation}

During training, routing is evaluated at the geometry-aware path point $\mathbf{z}_{k,\alpha}$ and global time $t_{k,\alpha}$. When source conditioning is used, the source endpoint $\mathbf{x}_k$ is also provided to the router and experts, following longitudinal flow-matching practice \citep{islam2025longitudinal}. Routing is trained with Straight-Through Gumbel-Softmax,
\begin{equation}
    w_m
    =
    \frac{
    \exp((\ell_m+\eta_m)/\tau_g)
    }{
    \sum_{q=1}^{M}\exp((\ell_q+\eta_q)/\tau_g)
    },
    \qquad
    \eta_m\sim\mathrm{Gumbel}(0,1),
\end{equation}
where $\tau_g$ is the router temperature. At inference, the candidate regime is the deterministic hard assignment
\begin{equation}
    \hat{r}(t,\mathbf{x})
    =
    \arg\max_m \ell_m(t,\mathbf{x}).
\end{equation}
Because $\hat{r}$ is derived from the velocity field, regime discovery is a decomposition of learned transport dynamics rather than a post hoc clustering of observations. Router regularizers discourage collapse and encourage sparse, temporally coherent assignments; definitions are provided in \Cref{app:penalties,app:composite_loss}.

\subsection{Training Objective}

Training proceeds in three stages: metric learning, bend-network training, and velocity-field training. In the final stage, the geometry model and bend network are frozen, and the velocity experts and router are trained on geometry-aware path points and tangents:
\begin{equation}
    \loss_{\mathrm{FM}}
    =
    \E
    \left[
    \left\|
    \vfield(t_{k,\alpha},\mathbf{z}_{k,\alpha})
    -
    \dot{\mathbf{z}}_{k,\alpha}
    \right\|_2^2
    \right].
\end{equation}
The full objective is
\begin{equation}
    \loss
    =
    \loss_{\mathrm{FM}}
    +
    \loss_{\mathrm{routing}}
    \label{eq:main_training_objective}
\end{equation}
where $\loss_{\mathrm{routing}}$ collects the router regularization terms. Hyperparameters and implementation details are reported in \Cref{app:pipeline,app:hyperparams,app:composite_loss}.

\section{Datasets and Baselines}
\subsection{Datasets}
We evaluate FLUX on four longitudinal snapshot datasets chosen to separate manifold transport from regime discovery. Stanford Bunny and Lorenz provide controlled systems with known geometry or known dynamical regimes \citep{turk1994zippered,lorenz1963deterministic}, whereas Widefield Calcium and Embryoid Body differentiation test whether the same framework recovers biologically meaningful transitions from high-dimensional population data \citep{cardin2020mesoscopic,ortegacaro2025selective,schiebinger2019optimal}.

\textbf{Stanford Bunny.} We use the Stanford Bunny mesh as a controlled manifold benchmark with known surface geometry \citep{turk1994zippered}. Eight ordered marginals are sampled along a geodesic path on the triangulated surface and embedded into $\mathbb{R}^{D}$ for $D\in\{3,5,10,20,50,100\}$ using an orthonormal projection. Marginals $\mu_1$ and $\mu_6$ are held out from velocity training and used only for evaluation. This tests whether longitudinal transport between unpaired snapshots remains on the manifold and interpolates to unseen marginal positions; full mesh construction, high-dimensional embedding, hold-out protocol, and surface-distance evaluation are given in Appendix~\ref{app:stanford_bunny}.

\textbf{Lorenz attractor.} We construct eight marginals from trajectory windows of the Lorenz system \citep{lorenz1963deterministic}. Marginals $\mu_0,\ldots,\mu_3$ are generated under the chaotic regime $(\sigma,\rho,\beta)=(10,28,8/3)$ and marginals $\mu_4,\ldots,\mu_7$ under the lower-$\rho$ regime $(10,12,8/3)$. The regime boundary is therefore known and occurs between $\mu_3$ and $\mu_4$. Each sample is a flattened $3\times20$ trajectory window, giving a 60-dimensional observation; simulation details are reported in Appendix~\ref{app:lorenz_generation}.

\textbf{Widefield calcium imaging.} We analyze widefield calcium imaging from 12 mice performing a Go/No-Go visual associative learning task \citep{cardin2020mesoscopic,ortegacaro2025selective}. Cortical activity is registered to 41 Allen-atlas areas, and each trial is represented as a 451-dimensional vector. Sessions are aligned into early, intermediate, and late learning phases using lick-index change points, yielding $T=22$ day-level marginals. Behavioral labels are used only for alignment and evaluation, not for velocity training or routing; preprocessing details are provided in Appendix~\ref{app:neural_preprocessing}.

\textbf{Embryoid body differentiation.} We use single-cell RNA-seq profiles across $T=5$ differentiation timepoints \citep{schiebinger2019optimal}. Cells are represented in PCA space \citep{pearson1901pca}, with up to $N=1{,}000$ cells sampled per marginal. Pluripotent, Commitment, and Differentiated stage labels are withheld during training and used only to evaluate regime discovery; preprocessing details are provided in Appendix~\ref{app:eb_preprocessing}.

\subsection{Baselines and Evaluation}
We compare FLUX against baselines that distinguish three modeling targets: adjacent-pair transport, full-chain longitudinal transport, and regime discovery. Adjacent-pair baselines include linear interpolation, static optimal transport (OT) \citep{peyre2019computational,cuturi2013sinkhorn}, and Independent Conditional Flow Matching (Independent CFM) \citep{lipman2023flow,tong2024improving}. These methods evaluate or learn transport between neighboring marginals independently and can obtain low adjacent-step Wasserstein distance without defining a single velocity field across the full sequence. Baseline implementation details are summarized in Appendix~\ref{app:baseline_notes}.

For longitudinal generative modeling, we compare against Marginal Matching Flow Matching (IMMFM), a Euclidean multi-marginal flow-matching baseline related to longitudinal and multi-marginal flow matching \citep{islam2025longitudinal,albergo2023multimarginal}. IMMFM defines full-chain ODE transport but does not decompose the velocity field into regimes. We also include FLUX without Manifold Learning, which preserves the same mixture-of-experts architecture \citep{shazeer2017outrageously,fedus2022switch}, router, Gumbel-Softmax estimator \citep{jang2017categorical}, and routing losses as FLUX but replaces geometry-aware conditional paths with Euclidean paths. This ablation tests whether regime recovery comes from expert capacity alone or from geometry-aware local dynamics. FLUX geometry, bend-network, routing, and regularization details are given in Appendices~\ref{app:penalties}--\ref{app:composite_loss}, with hyperparameters in Appendix~\ref{app:hyperparams}.

For regime discovery, we compare router assignments against Gaussian, K-means \citep{macqueen1967kmeans}, Gaussian-mixture/EM \citep{dempster1977em}, PCA+K-means \citep{pearson1901pca,macqueen1967kmeans}, spectral clustering \citep{ng2002spectral}, HMM \citep{rabiner1989tutorial}, rSLDS \citep{linderman2017rslds}, and SRNN \citep{zhang2024srnn} baselines using the same segment-level labels across methods. Transport is evaluated with adjacent-step Wasserstein distance ($\mathrm{WD}_1$), two-hop Wasserstein distance when applicable, and full-chain Wasserstein distance ($\mathrm{WD}_{\mathrm{fc}}$). The full-chain metric integrates the learned ODE \citep{chen2018neural} from the initial marginal through the longitudinal sequence and is the stricter test of whether a single continuous process is consistent with all observed snapshots. Regime recovery is evaluated with segment-level adjusted Rand index (ARI) and normalized mutual information (NMI). On the Stanford Bunny, we additionally report held-out Wasserstein distance after projection back to 3D. Metric definitions and evaluation protocols are given in Appendices~\ref{app:evaluation_protocol} and~\ref{app:evaluation_metrics}.

\vspace{-0.3cm}
\section{Results}
\subsection{Geometry-Aware Paths Preserve Surface Transport on the Stanford Bunny}
\label{sec:results_bunny}

\begin{figure}[t]
    \centering
    \includegraphics[width=0.8\textwidth]{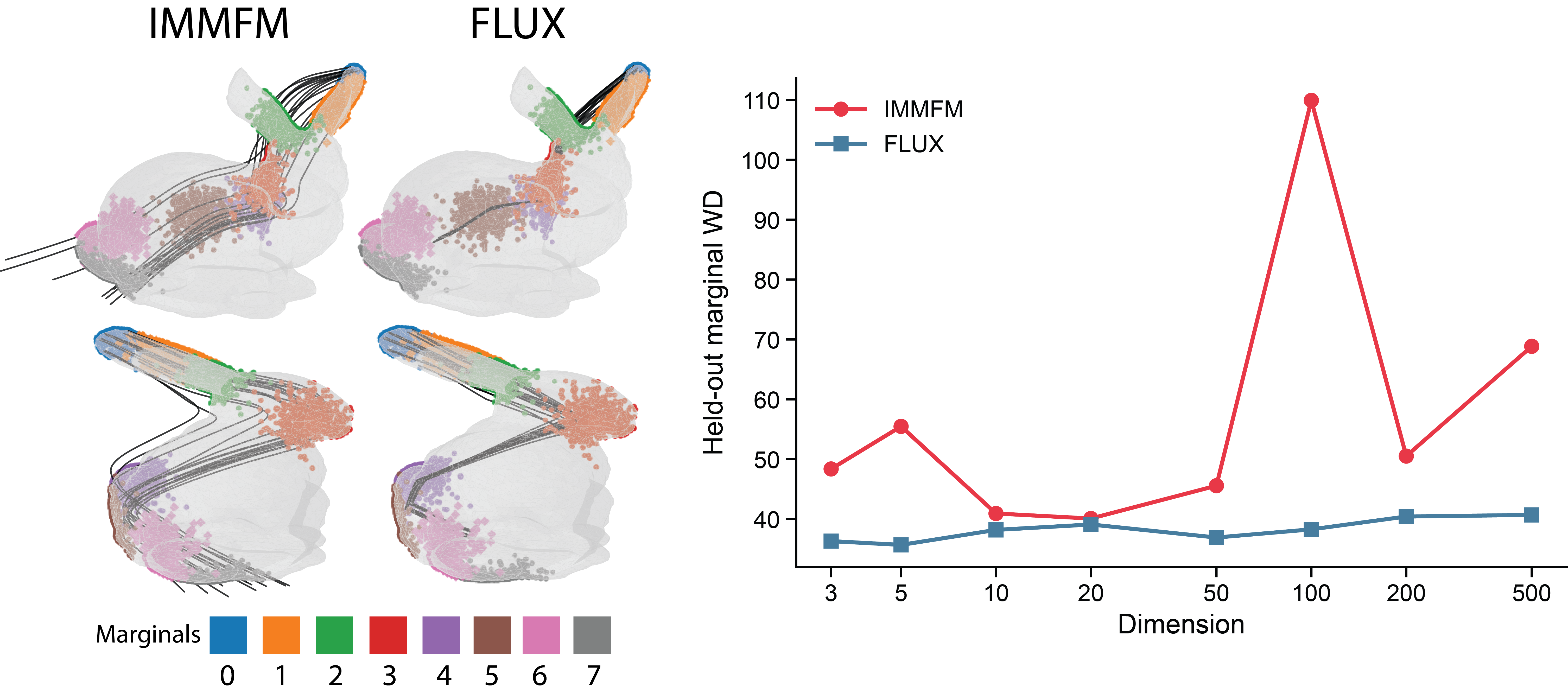}%
    \caption{\textbf{Stanford Bunny dimensionality ablation.} Eight ordered marginals are sampled on a geodesic path over the Stanford Bunny mesh, with two intermediate marginals held out from velocity training. The same surface is embedded into $\mathbb{R}^{D}$ for $D\in\{3,5,10,20,50,100\}$. Evaluation integrates each model sequentially through all eight marginal times and reports held-out Wasserstein distance.}
    \vspace{-0.5cm}
    \label{fig:bunny_dim_sweep}
\end{figure}

We first test the geometric premise of FLUX: longitudinal transport between unpaired snapshots should follow the data manifold rather than Euclidean shortcuts through the ambient space. This is a central issue in biological snapshot systems, where only sparse labels may be available, but the full collection of observations can still define the geometry of the state space. To isolate this problem, we use the Stanford Bunny mesh as a known three-dimensional manifold and sample eight ordered empirical marginals along a geodesic path on its surface, holding out two intermediate marginals (second and sixth marginals) from velocity training.

Figure~\ref{fig:bunny_dim_sweep} evaluates whether geometry-aware paths preserve this surface transport and the effect of ambient dimension to the fit. The 3D bunny surface is embedded into $\mathbb{R}^{D}$ for $D\in\{3,5,10,20,50,100\}$ using an orthonormal random projection, preserving distances within the embedded mesh while increasing co-dimension. Both IMMFM and FLUX are integrated sequentially through all eight marginal times and evaluated by coordinate-wise Wasserstein distance after projection back to 3D and by mean distance to the mesh surface. Held-out Wasserstein distance measures interpolation to unseen marginals, while surface deviation measures whether generated samples remain on the valid manifold. This experiment tests whether manifold geometry learned from available observations can improve longitudinal transport when the observed space is high-dimensional but the intrinsic dynamics are low-dimensional.

\begin{figure}[t]
    \centering
    \includegraphics[width=\textwidth]{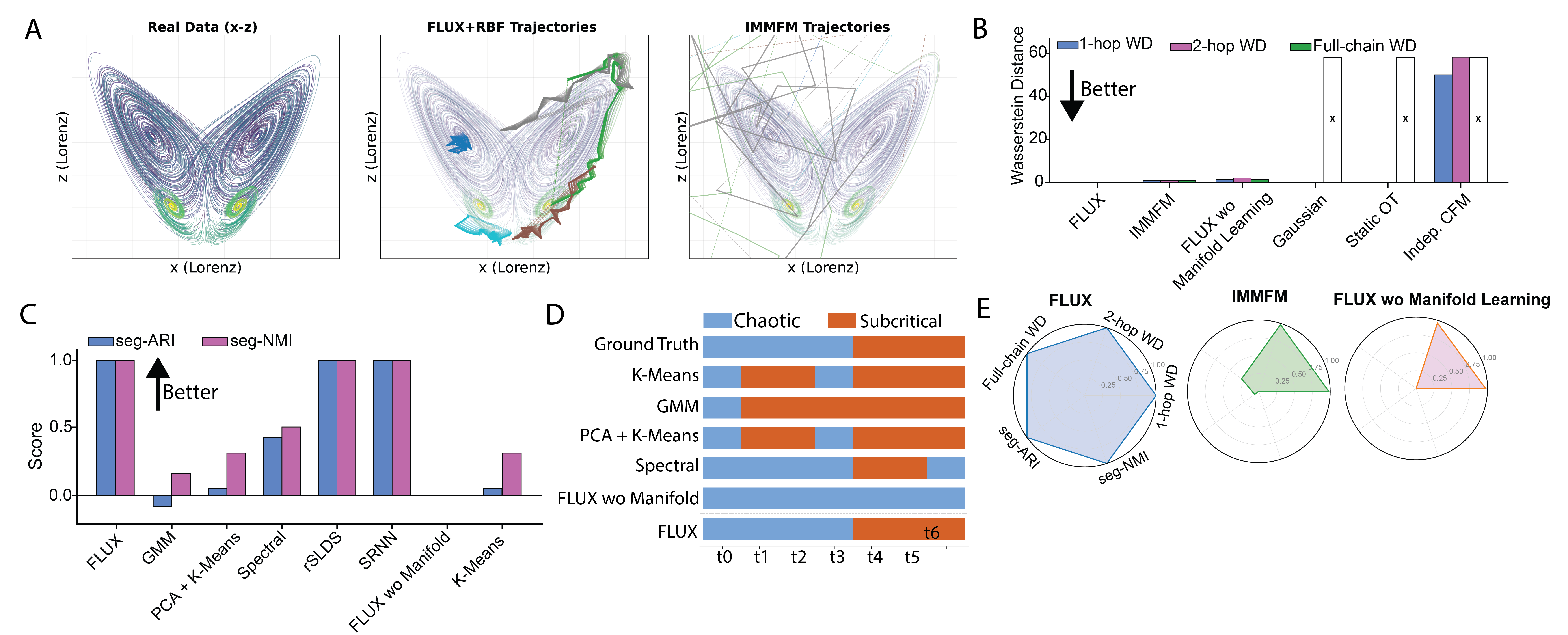}
    \vspace{-0.4cm}
    \caption{\textbf{Lorenz dynamical-system benchmark.} \textbf{(A)} Two-dimensional visualization of trajectory-window samples colored by the ground-truth Lorenz parameter regime. Each model input is a flattened $3\times20$ trajectory segment. \textbf{(B)} Generative transport metrics: one-hop Wasserstein distance, two-hop Wasserstein distance, and full-chain Wasserstein distance. \textbf{(C)} Segment-level regime-discovery metrics, including adjusted Rand index (ARI) and normalized mutual information (NMI). \textbf{(D)} Temporal regime assignments predicted by each method compared with the ground-truth parameter switch. \textbf{(E)} Radar summary of transport and regime-discovery metrics for the main longitudinal models.}
    \label{fig:lorenz}
    \vspace{-0.75cm}
\end{figure}

\subsection{FLUX Recovers Parameter Regimes in Lorenz Dynamics}

We next test whether geometry-aware transport is necessary for discovering regimes encoded in local dynamics rather than static spatial clusters. The Lorenz benchmark provides a controlled setting with a known dynamical transition: marginals $\mu_0,\ldots,\mu_3$ are sampled from trajectory windows generated under the chaotic parameter regime $(\sigma,\rho,\beta)=(10,28,8/3)$, whereas marginals $\mu_4,\ldots,\mu_7$ are sampled under the lower-$\rho$ regime $(10,12,8/3)$. The ground-truth regime boundary is therefore the parameter switch between $\mu_3$ and $\mu_4$.

IMMFM achieves the lowest Wasserstein distances, indicating strong Euclidean multi-marginal transport (Table~\ref{tab:lorenz}, Figure~\ref{fig:lorenz}D). However, the scientific target here is not only endpoint matching, but recovery of the latent change in dynamics. FLUX trades higher transport error for exact regime recovery: router $g$ identifies the chaotic/subcritical boundary with segment-level adjusted Rand index (seg-ARI) $=1.0$ and segment-level normalized mutual information (seg-NMI) $=1.0$ (Table~\ref{tab:regime}). FLUX without Manifold Learning collapses to a single expert despite using the same MoE architecture and routing objective, showing that expert capacity alone is insufficient. Geometry-aware paths expose the curvature and velocity signatures needed for expert specialization when regime identity is encoded in local dynamics.

\begin{figure}[t]
    \centering
    \includegraphics[width=0.8\textwidth]{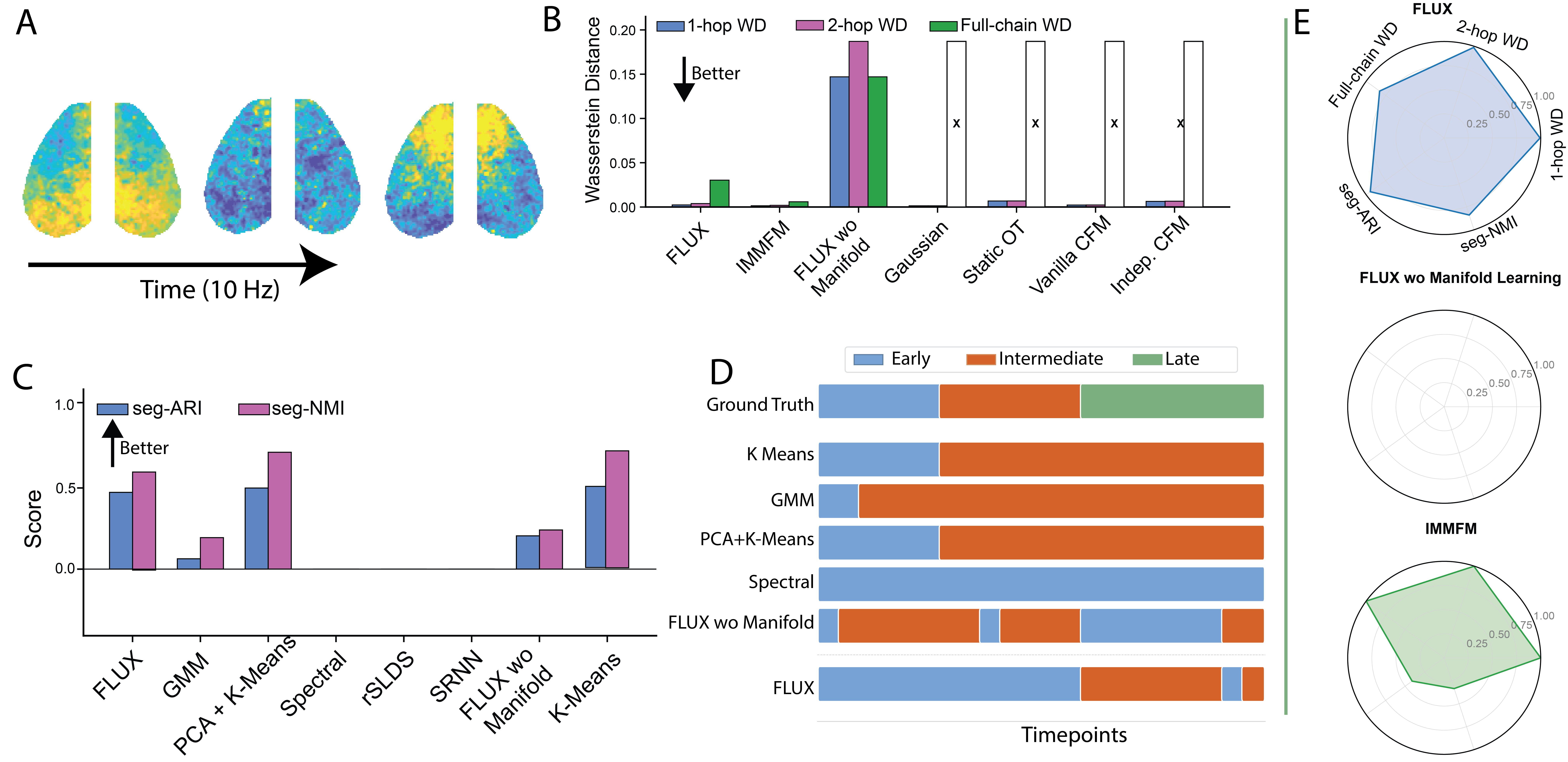}
    \caption{\textbf{Widefield calcium-imaging benchmark.} \textbf{(A)} Example post-stimulus cortical activity represented as brain area by time. Each trial is flattened into a 451-dimensional vector. \textbf{(B)} Generative transport metrics across Widefield Calcium baselines. \textbf{(C)} Segment-level regime-discovery metrics. \textbf{(D)} Temporal expert assignments compared with early, intermediate, and late behavioral learning labels. \textbf{(E)} Radar summary of transport and regime metrics.}
    \label{fig:neural}
\end{figure}

\subsection{FLUX Identifies Learning-Related Regimes in Neural Population Dynamics}

Next, we asked if the same principle reveals latent biological transitions during learning. In longitudinal neural recordings, animals learn as a function of individual trials over multiple days. Single trials are represented as cortical population activity over brain regions and time, and marginals correspond to individual sessions ordered according to learning stage. The biological question is whether changes in neural population dynamics reveal a transition from early generalized stimulus--reward expectation to late stimulus-specific discrimination, without providing behavioral labels during training.

FLUX substantially reduces full-chain error relative to single-velocity Vanilla CFM, indicating that manifold learning and expert decomposition improves long-horizon transport through the learning trajectory (Table~\ref{tab:neural_table}). Although FLUX without Manifold Learning achieves the lowest adjacent-step Wasserstein distance, it discovers weaker regime structure, again separating local transport accuracy from dynamical regime recovery. Router $g$ reaches seg-ARI$=0.461$ and seg-NMI$=0.589$ (Table~\ref{tab:regime}), outperforming Euclidean routing and switching dynamical baselines while approaching the best clustering baseline. The inferred boundary separates early/intermediate from late learning, consistent with the emergence of behavioral discrimination between CS$+$ and CS$-$ stimuli. This suggests that cortical population dynamics reorganize into distinct dynamical regimes with associative learning.

\begin{figure}[t]
    \centering
    \includegraphics[width=0.8\textwidth]{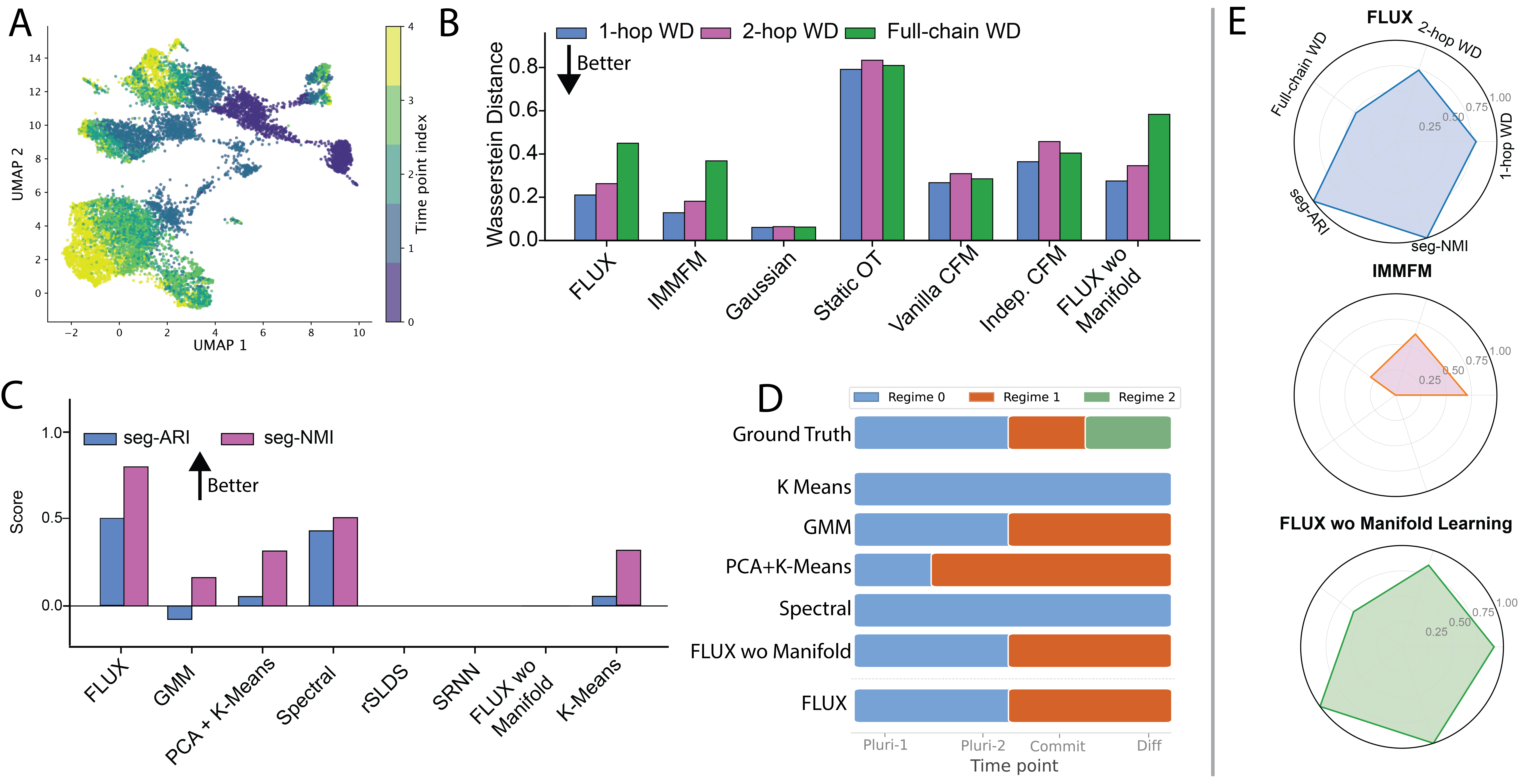}
    \caption{\textbf{Embryoid body differentiation benchmark.} \textbf{(A)} RNA-seq profiles projected into UMAP space and colored by differentiation stage. Each model input is a PCA representation of gene expression. \textbf{(B)} Generative transport metrics. \textbf{(C)} Segment-level regime-discovery metrics. \textbf{(D)} Temporal expert assignments compared with pluripotent, commitment, and differentiated stage labels. \textbf{(E)} Radar summary of transport and regime metrics.}
    \label{fig:eb}
    \vspace{-0.7cm}
\end{figure}

\subsection{FLUX Recovers Differentiation Stages When Stages Are Spatially Separable}

Finally, we test whether regime recovery depends on geometry-aware local dynamics in a biological system where stages are more spatially separated. In embryoid body differentiation, marginals follow developmental time, and the expected transition is from pluripotent populations toward more differentiated gene-expression programs. This provides a complementary case to Lorenz and neural learning: differentiation stages can be partially separated in representation space, so static geometry and Euclidean routing may already contain substantial regime information.

 Linear interpolation attains near-zero adjacent-step WD because it directly evaluates endpoint interpolation under random pairing, but it does not define a learned shared velocity field or full-chain ODE transport. Among learned longitudinal flow models, FLUX without Manifold Learning gives the lowest adjacent-step transport error. For regime discovery, FLUX and FLUX without Manifold Learning both match the best non-neural baseline, reaching seg-ARI$=0.571$ and seg-NMI$=0.800$ (Table~\ref{tab:regime}, Figure\ref{fig:eb}E). This contrast supports the main interpretation: Euclidean routing can succeed when regimes are spatially clustered, but geometry-aware routing is most important when regimes are encoded in local dynamics rather than simple separation in state space.

\section{Discussion}

We introduced FLUX, a geometry-aware longitudinal flow-matching framework that combines learned manifold-aware paths with a mixture-of-experts velocity field. FLUX is designed for unpaired biological snapshots, where the goal is not only to move mass between timepoints but also to identify changes in the dynamics that generate the sequence.

Our results show that geometry matters when sparse longitudinal labels are embedded in a richer unlabeled state space. The Stanford Bunny control demonstrates this in a setting with a known manifold: metric learning improves transport along curved surfaces, including under high-dimensional embeddings. This supports the central motivation for neural and genomic data, where observations are high-dimensional but structured by lower-dimensional latent manifolds.

The results also show that transport accuracy and regime discovery are distinct. IMMFM can achieve strong Wasserstein performance, but it does not expose changes in the velocity field. FLUX instead trades purely local transport optimization for interpretable dynamical structure. This is most important when regimes are defined by local dynamics rather than obvious spatial separation, as seen in Lorenz and the neural datasets.

Biologically, the inferred regimes should be interpreted as candidate transitions in population dynamics rather than clusters of observations. In neural data, FLUX recovers a transition aligned with the emergence of CS$+$ versus CS$-$ behavioral discrimination. In embryoid body differentiation, it identifies regimes consistent with progression from pluripotent toward differentiated states.

Several limitations remain. FLUX is more computationally expensive than Euclidean longitudinal flow matching because it trains 3 different networks. The number of experts is also fixed in advance; adaptive routing, could make regime discovery less dependent on this choice. Furthermore more, when local dynamics are noisy or weakly sampled, inferred regimes should be interpreted cautiously.

Together, these results shows FLUX is a  geometry-aware trajectory inference for longitudinal regime discovery. By applying learned metric paths across a marginal chain and decomposing the resulting velocity field into sparse regimes, FLUX provides a framework for interpretable dynamical modeling of unpaired biological snapshots. Future work should focus on cheaper geometry learning, adaptive regime discovery, and stronger biological validation.

================================================================

\bibliographystyle{plainnat}
\bibliography{references}

\newpage
\appendix
\appendix

\section{Three-Stage Training Pipeline}
\label{app:pipeline}

FLUX is trained in three sequential stages. First, a geometry backend is trained on pooled samples from all marginals to define a data-dependent metric $\mathbf{G}(\mathbf{x})$. Second, a bend network $B_\psi$ is trained with the frozen metric to produce geometry-aware interpolants between adjacent-marginal endpoints. Third, the velocity model---either a single VelocityNet or an MoE MixtureVelocityNet---is trained using bend-interpolated samples and conditional velocity targets.

All stages use AdamW with dataset-specific learning rates and weight decay. The bend stage uses ReduceLROnPlateau scheduling with factor 0.5 and patience 10; the velocity stage uses a static learning rate. Hyperparameters are reported in \Cref{tab:hyperparams}. The staged design prevents unstable gradients from the metric-learning and velocity objectives from interfering during optimization, while allowing each module to be tuned independently.

\subsection{RBF-MLP Metric for Higher-Dimensional Observations}
\label{app:rbf_mlp_metric}

For observations with ambient dimension $d\geq 5$, we use an RBF-MLP metric rather than applying the RBF score directly in the observed coordinate system. The goal is to retain the same distance-to-manifold logic as the low-dimensional RBF backend while avoiding the poor distance concentration and noisy kernel neighborhoods that occur in high-dimensional spaces. The backend first learns a feature map
\begin{equation}
    q_\theta:\R^d\rightarrow\R^{d_z},
\end{equation}
where $d_z$ is a compact metric-feature dimension. Let $\mathbf{z}=q_\theta(\mathbf{x})$. A set of $M$ kernel centers $\{\mathbf{c}_m\}_{m=1}^M$ is initialized from pooled training samples in feature space and updated during metric training. The deep-kernel manifold score is
\begin{equation}
    h_\theta(\mathbf{x})
    =
    \sigma\left(
    a\,\log\sum_{m=1}^{M}
    \exp\left[-\frac{\|q_\theta(\mathbf{x})-\mathbf{c}_m\|_2^2}{2\sigma_k^2}\right]
    + b
    \right),
    \label{eq:app_rbf_mlp_score}
\end{equation}
where $\sigma(\cdot)$ is the logistic function, $\sigma_k$ is the kernel bandwidth, and $a,b$ are learned scalar calibration parameters. The induced scalar metric is
\begin{equation}
    M_{\mathrm{RBF\text{-}MLP}}(\mathbf{x})
    =
    \left(h_\theta(\mathbf{x})+\epsilon\right)^{-\alpha},
\end{equation}
and the bend network uses the isotropic metric
\begin{equation}
    \mathbf{G}_{\mathrm{RBF\text{-}MLP}}(\mathbf{x})
    =
    M_{\mathrm{RBF\text{-}MLP}}(\mathbf{x})\mathbf{I}.
\end{equation}
Thus, both geometry backends used in this paper share the same interpretation: points estimated to be far from the data manifold receive high metric cost. The difference is that the direct RBF metric is used when $d<5$, whereas RBF-MLP learns the kernel space before estimating the RBF manifold score in higher-dimensional datasets such as Lorenz trajectory windows, NeuralTable, and embryoid body expression states.

\subsection{Bend Network}
\label{app:bend_network}

The bend network takes as input
\begin{equation}
    [\mathbf{x}_0;\mathbf{x}_1;\tau]\in\R^{2d+1}
\end{equation}
and outputs a displacement $\Delta_\psi\in\R^d$. The interpolant is
\begin{equation}
    B_\psi(\mathbf{x}_0,\mathbf{x}_1,\tau)
    =
    (1-\tau)\mathbf{x}_0+\tau\mathbf{x}_1
    +
    \gamma(\tau)\Delta_\psi(\mathbf{x}_0,\mathbf{x}_1,\tau),
    \qquad
    \gamma(\tau)=4\tau(1-\tau).
    \label{eq:app_bend_network}
\end{equation}
The factor $\gamma(\tau)$ forces the bend correction to vanish at $\tau=0$ and $\tau=1$, preserving endpoint fidelity. The corresponding velocity target is computed by differentiating the bend path with respect to local interpolation time:
\begin{equation}
    \mathbf{u}_{\tau}
    =
    \partial_\tau B_\psi(\mathbf{x}_0,\mathbf{x}_1,\tau).
\end{equation}
The bend network is a 2-layer MLP with ReLU activations. Hidden dimensions are dataset-specific and listed in \Cref{tab:hyperparams}.

\section{Hyperparameter Details}
\label{app:hyperparams}

\begin{table}[h]
\centering
\caption{Key hyperparameters across experiments. ``Single'' and ``per-expert'' hidden dimensions are listed separately where they differ.}
\label{tab:hyperparams}
\resizebox{\textwidth}{!}{
\begin{tabular}{lccc}
\toprule
\textbf{Parameter} & \textbf{Lorenz} & \textbf{NeuralTable} & \textbf{Embryoid Body} \\
\midrule
\multicolumn{4}{l}{\textit{Architecture}} \\
Hidden dim, single velocity & 64 & 256, 128 & 128 \\
Hidden dim, per expert & 8 & 256, 128 & 64 \\
Velocity layers & 2 & 2 & 2 \\
Time embedding dim & 16 & 16 & 32 \\
Number of experts $K$ & 2 & 3 & 3 \\
\midrule
\multicolumn{4}{l}{\textit{Training}} \\
Geometry epochs & 50 & 50 & 100 \\
Bend epochs & 100 & 100 & 150 \\
Velocity epochs & 180 & 400 & 300 \\
Batch size & 32 & 32 & 64 \\
Learning rate, geometry/bend & $10^{-4}$ & $5\times10^{-5}$ & $10^{-3}$ \\
Learning rate, velocity & $10^{-4}$ & $5\times10^{-6}$ & $10^{-6}$ \\
Weight decay & $10^{-5}$ & $10^{-5}$ & $10^{-5}$ \\
Early stopping patience & 30 & 60 & 60 \\
\midrule
\multicolumn{4}{l}{\textit{Gumbel-Softmax}} \\
$\tau_{\mathrm{init}}$ & 1.0 & 1.0 & 1.0 \\
$\tau_{\mathrm{min}}$ & 0.2 & 0.2 & 0.05 \\
$\tau_{\mathrm{decay}}$ & 0.9995 & 0.9995 & 0.99 \\
Soft epochs & 50 & 50 & 50 \\
\midrule
\multicolumn{4}{l}{\textit{Geometry backend}} \\
Metric type & RBF-MLP & RBF-MLP & RBF-MLP \\
RBF-MLP feature dim $d_z$ & 32 & 32 & 32 \\
Number of RBF centers $M$ & 128 & 256 & 256 \\
\midrule
\multicolumn{4}{l}{\textit{Gating penalties}} \\
Load-balance weight & $0.1\rightarrow0.001$ & $0.1\rightarrow0.001$ & $0.1\rightarrow0.001$ \\
Z-loss weight & 0.01 & 0.01 & 0.01 \\
Clustering weight & 0.5 & 0.5 & 0.5 \\
Temporal TV weight & 0.05 & 0.05 & 0.05 \\
Confidence weight & 0.05 & 0.05 & 0.05 \\
Diversity weight & 1.0 & 1.0 & 1.0 \\
Velocity regularization weight & 0.10 & 0.05 & 0.02 \\
\bottomrule
\end{tabular}
}
\end{table}

\subsection{Evaluation Protocol}
\label{app:evaluation_protocol}

We evaluate two complementary properties: longitudinal transport and unsupervised regime discovery. Transport quality is measured with sliced Wasserstein distance after one-step, two-step, and full-chain transport:
\begin{align}
    \mathrm{WD}_1 &=
    \frac{1}{T-1}\sum_{k=0}^{T-2}
    W_1\left((\phi_{t_k\rightarrow t_{k+1}})_{\#}\mu_k,\mu_{k+1}\right),\\
    \mathrm{WD}_2 &=
    \frac{1}{|\mathcal{P}_2|}
    \sum_{(i,j)\in\mathcal{P}_2}
    W_1\left((\phi_{t_i\rightarrow t_j})_{\#}\mu_i,\mu_j\right),
    \qquad
    \mathcal{P}_2=\{(i,i+2):0\leq i\leq T-3\},\\
    \mathrm{WD}_{\mathrm{fc}} &=
    W_1\left((\phi_{t_0\rightarrow t_{T-1}})_{\#}\mu_0,\mu_{T-1}\right).
\end{align}
Here $W_1$ is estimated with random one-dimensional projections. $\mathrm{WD}_1$ measures adjacent-step accuracy, $\mathrm{WD}_2$ measures short multi-step consistency, and $\mathrm{WD}_{\mathrm{fc}}$ measures accumulated error after integrating from the first to the final marginal. Static OT, linear interpolation, and independent pairwise models can be strong adjacent-step baselines, but they do not always define a single shared full-chain ODE transport; when they do not, $\mathrm{WD}_{\mathrm{fc}}$ is not reported.

Regime discovery is evaluated at the temporal-segment level. Let $\mathbf{z}=(z_1,\ldots,z_S)$ denote the ground-truth regime labels for $S$ temporal segments, and let $\hat{\mathbf{z}}=(\hat{z}_1,\ldots,\hat{z}_S)$ denote the predicted segment labels. For FLUX, $\hat{z}_s$ is the majority hard expert assignment $\arg\max_j w_j(t,\mathbf{x})$ among samples in segment $s$. For clustering baselines, $\hat{z}_s$ is the majority cluster label among samples in segment $s$. This segment-level evaluation asks whether a method recovers temporal regime boundaries rather than whether it classifies each individual sample.

The adjusted Rand index (ARI) compares the pairwise agreement between $\mathbf{z}$ and $\hat{\mathbf{z}}$ while correcting for chance. Let $n_{ab}$ be the contingency-table count of segments assigned to true regime $a$ and predicted regime $b$, let $n_{a\cdot}=\sum_b n_{ab}$, $n_{\cdot b}=\sum_a n_{ab}$, and let $N=\sum_{a,b}n_{ab}$ be the number of evaluated segments. Then
\begin{equation}
    \mathrm{ARI}
    =
    \frac{
    \sum_{a,b}\binom{n_{ab}}{2}
    -
    \frac{\sum_a\binom{n_{a\cdot}}{2}\sum_b\binom{n_{\cdot b}}{2}}{\binom{N}{2}}
    }{
    \frac{1}{2}\left[
    \sum_a\binom{n_{a\cdot}}{2}+\sum_b\binom{n_{\cdot b}}{2}
    \right]
    -
    \frac{\sum_a\binom{n_{a\cdot}}{2}\sum_b\binom{n_{\cdot b}}{2}}{\binom{N}{2}}
    }.
\end{equation}
ARI equals $1$ for perfect agreement and has expected value near $0$ under random assignments; negative values indicate worse-than-chance agreement.

Normalized mutual information (NMI) measures the shared information between true and predicted segment labels. With empirical probabilities $p_{ab}=n_{ab}/N$, $p_{a\cdot}=n_{a\cdot}/N$, and $p_{\cdot b}=n_{\cdot b}/N$, the mutual information is
\begin{equation}
    I(\mathbf{z};\hat{\mathbf{z}})
    =
    \sum_{a,b}p_{ab}
    \log\frac{p_{ab}}{p_{a\cdot}p_{\cdot b}},
\end{equation}
with zero-contribution convention for $p_{ab}=0$. We report
\begin{equation}
    \mathrm{NMI}
    =
    \frac{2I(\mathbf{z};\hat{\mathbf{z}})}
    {H(\mathbf{z})+H(\hat{\mathbf{z}})},
\end{equation}
where $H(\cdot)$ denotes entropy. NMI lies in $[0,1]$, with $1$ indicating perfect agreement.

We also report gating entropy and switch rate. Gating entropy is the average entropy of the router probabilities,
\begin{equation}
    H_{\mathrm{gate}}
    =
    -\frac{1}{B}\sum_{i=1}^{B}\sum_{j=1}^{K}w_{ij}\log(w_{ij}+\epsilon),
\end{equation}
where $B$ is the number of evaluated samples and $K$ is the number of experts. Switch rate is the fraction of adjacent temporal segments whose majority predicted expert label changes. These metrics are interpreted jointly: low transport error indicates generative fidelity, while high ARI/NMI indicates recovery of the temporal regime structure.

\section{Gating Penalty Details}
\label{app:penalties}

Let $g$ denote the router network, and let $\mathbf{w}_i\in\Delta^{K-1}$ denote the routing weights produced for sample $i$, where $K$ is the number of experts. Let $\bar{\mathbf{w}}=\frac{1}{B}\sum_{i=1}^{B}\mathbf{w}_i$ be the batch-averaged routing distribution for a minibatch of size $B$, and let $\boldsymbol{\ell}_i=g(t_i,\mathbf{x}_i)\in\R^K$ denote the router logits. The routing penalties are designed to prevent expert collapse, encourage discrete assignments, and promote temporal coherence.

\subsection{Diversity and Sparsity}
\label{app:diversity_sparsity}

Batch-level diversity prevents all inputs from being routed to the same expert:
\begin{equation}
    \loss_{\mathrm{div}}
    =
    1-\frac{H(\bar{\mathbf{w}})}{\log K},
    \label{eq:app_diversity}
\end{equation}
where $H(\cdot)$ denotes entropy. When all samples are routed to a single expert, $H(\bar{\mathbf{w}})\rightarrow 0$ and the penalty is large.

Per-sample sparsity encourages individual samples to commit to a single expert:
\begin{equation}
    \loss_{\mathrm{sp}}
    =
    \frac{1}{B}
    \sum_{i=1}^{B}
    \frac{H(\mathbf{w}_i)}{\log K}.
    \label{eq:app_sparsity}
\end{equation}
Diversity and sparsity can be simultaneously satisfied when each sample uses a sharp expert assignment but the batch uses multiple experts.

\subsection{Temporal Consistency}
\label{app:temporal_consistency}

To discourage pathological oscillations in expert assignment, samples are grouped by nearby flow times. Let $\mathcal{G}$ denote the set of time-proximity groups. The consistency penalty is
\begin{equation}
    \loss_{\mathrm{con}}
    =
    \frac{1}{|\mathcal{G}|}
    \sum_{q\in\mathcal{G}}
    \mathrm{Var}
    \left[
    \mathbf{w}_i : i\in q
    \right].
    \label{eq:app_consistency}
\end{equation}
Here, $q$ indexes a temporal group. We avoid using $g$ as a group index because $g$ denotes the router network.

\subsection{Load Balance, Confidence, and Clustering}
\label{app:extended_penalties}

For all Gumbel-enabled configurations, we use an extended penalty suite. The load-balance loss follows sparse MoE practice:
\begin{equation}
    \loss_{\mathrm{lb}}
    =
    K\sum_{j=1}^{K}F_jP_j,
    \qquad
    F_j=
    \frac{1}{B}
    \sum_{i=1}^{B}
    \mathbf{1}[\arg\max_m w_{im}=j],
    \qquad
    P_j=
    \frac{1}{B}
    \sum_{i=1}^{B}
    w_{ij}.
    \label{eq:app_load_balance}
\end{equation}
Here, $F_j$ is the empirical fraction of hard assignments to expert $j$, and $P_j$ is the mean soft usage of expert $j$.

The router z-loss penalizes saturated logits:
\begin{equation}
    \loss_z
    =
    \frac{1}{B}
    \sum_{i=1}^{B}
    \left(
    \log\sum_{j=1}^{K}e^{\ell_{ij}}
    \right)^2.
    \label{eq:app_z_loss}
\end{equation}

The confidence penalty encourages a high maximum routing probability:
\begin{equation}
    \loss_{\mathrm{conf}}
    =
    \frac{1}{B}
    \sum_{i=1}^{B}
    \left(1-\max_j w_{ij}\right).
    \label{eq:app_confidence}
\end{equation}

The clustering penalty uses DEC-style self-sharpening. Let
\begin{equation}
    Q_{ij}
    =
    \mathrm{softmax}(\boldsymbol{\ell}_i)_j
\end{equation}
be the soft assignment probability before Gumbel sampling. The sharpened target distribution is
\begin{equation}
    P_{ij}
    =
    \frac{Q_{ij}^2/\sum_i Q_{ij}}
    {\sum_{m=1}^{K}Q_{im}^2/\sum_i Q_{im}},
\end{equation}
and the clustering loss is
\begin{equation}
    \loss_{\mathrm{clust}}
    =
    \mathrm{KL}(Q\,\|\,P).
    \label{eq:app_clustering}
\end{equation}

\subsection{Temporal Smoothness and Segment-Level Penalties}
\label{app:temporal_smoothness}

For datasets with explicit temporal segments, let $\mathcal{S}$ denote the set of segments and let $\bar{\mathbf{w}}_s$ be the mean routing vector within segment $s$. The segment-consistency penalty is
\begin{equation}
    \loss_{\mathrm{seg\mbox{-}con}}
    =
    \frac{1}{|\mathcal{S}|}
    \sum_{s\in\mathcal{S}}
    \mathrm{Var}
    \left[
    \mathbf{w}_i : i\in s
    \right].
\end{equation}
The segment-sharpness penalty is
\begin{equation}
    \loss_{\mathrm{seg\mbox{-}sharp}}
    =
    \frac{1}{|\mathcal{S}|}
    \sum_{s\in\mathcal{S}}
    \frac{H(\bar{\mathbf{w}}_s)}{\log K}.
\end{equation}
A temporal total-variation penalty encourages smooth changes across adjacent segments:
\begin{equation}
    \loss_{\mathrm{TV}}
    =
    \frac{1}{|\mathcal{S}|-1}
    \sum_{s=1}^{|\mathcal{S}|-1}
    \left\|
    \bar{\mathbf{w}}_s-\bar{\mathbf{w}}_{s+1}
    \right\|_1.
\end{equation}
Finally, a contiguity penalty discourages individual experts from activating in temporally scattered intervals:
\begin{equation}
    \loss_{\mathrm{contig}}
    =
    \frac{1}{K}
    \sum_{j=1}^{K}
    \frac{
    \mathrm{Var}_s
    \left[
    s\cdot\bar{w}_{s,j}
    \right]
    }{
    \sum_s \bar{w}_{s,j}+\epsilon
    }.
\end{equation}

\section{Composite Training Objective}
\label{app:composite_loss}

The velocity-stage objective combines geometry-aware flow matching, velocity regularization, parameter regularization, and routing penalties:
\begin{equation}
    \loss
    =
    \loss_{\mathrm{FM}}
    +
    \lambda_{\mathrm{L2}}\loss_{\mathrm{L2}}
    +
    \lambda_{\mathrm{vel}}\loss_{\mathrm{vel}}
    +
    \loss_{\mathrm{routing}}.
    \label{eq:app_composite_objective}
\end{equation}
The flow-matching term is
\begin{equation}
    \loss_{\mathrm{FM}}
    =
    \E_{k,\tau,(\mathbf{x}_0,\mathbf{x}_1)}
    \left[
    \left\|
    \vfield(t,\mathbf{x}_{\tau})-\mathbf{u}_{\tau}
    \right\|_2^2
    \right],
    \label{eq:app_fm_loss}
\end{equation}
where $t=(k+\tau)/(T-1)$, $\mathbf{x}_{\tau}=B_\psi(\mathbf{x}_0,\mathbf{x}_1,\tau;\mathbf{G})$, and $\mathbf{u}_{\tau}=\partial_\tau B_\psi(\mathbf{x}_0,\mathbf{x}_1,\tau;\mathbf{G})$. The velocity regularizer is
\begin{equation}
    \loss_{\mathrm{vel}}
    =
    \E_{t,\mathbf{x}}
    \left[
    \|\vfield(t,\mathbf{x})\|_2^2
    \right],
\end{equation}
and $\loss_{\mathrm{L2}}$ is parameter-level weight decay.

For MoE models, the router $g$ is optimized through the routing objective
\begin{align}
    \loss_{\mathrm{routing}}
    =
    &
    \lambda_{\mathrm{div}}\loss_{\mathrm{div}}
    +
    \lambda_{\mathrm{con}}\loss_{\mathrm{con}}
    +
    \lambda_{\mathrm{sp}}\loss_{\mathrm{sp}}
    +
    \lambda_{\mathrm{lb}}\loss_{\mathrm{lb}}
    +
    \lambda_z\loss_z
    +
    \lambda_{\mathrm{conf}}\loss_{\mathrm{conf}}
    +
    \lambda_{\mathrm{clust}}\loss_{\mathrm{clust}}
    \nonumber\\
    &
    +
    \lambda_{\mathrm{seg\mbox{-}con}}\loss_{\mathrm{seg\mbox{-}con}}
    +
    \lambda_{\mathrm{seg\mbox{-}sharp}}\loss_{\mathrm{seg\mbox{-}sharp}}
    +
    \lambda_{\mathrm{TV}}\loss_{\mathrm{TV}}
    +
    \lambda_{\mathrm{contig}}\loss_{\mathrm{contig}}.
    \label{eq:app_routing_loss}
\end{align}
Only the subset of active penalties is used for each dataset. For Lorenz, NeuralTable, and Embryoid Body, the extended penalty suite is active with dataset-specific weights listed in \Cref{tab:hyperparams}.

The data-fit term uses a finite-safe MSE: non-finite predicted or target velocities are replaced with zero, residuals are clamped to $[-10^3,10^3]$, and the MSE is computed only over finite entries. This prevents occasional numerical failures from upstream geometry computations from propagating through the full batch loss.

\section{Lorenz Data Generation}
\label{app:lorenz_generation}

The Lorenz benchmark is designed as an IMMFM-style multi-marginal dataset with a known regime transition. It contains $T=8$ discrete empirical marginals, indexed by $k=0,\ldots,7$, with normalized marginal times given by $\mathrm{linspace}(0,1,8)$. Training uses adjacent-marginal transports $(\mu_k,\mu_{k+1})$. The eight marginals are split into two contiguous regime blocks: marginals $\mu_0,\ldots,\mu_3$ belong to regime 0, and marginals $\mu_4,\ldots,\mu_7$ belong to regime 1. Thus, the dataset contains eight marginal timepoints but only two Lorenz parameter regimes.

The continuous dynamics are generated from the standard Lorenz system
\begin{align}
    \frac{dx}{dt} &= \sigma(y-x),\\
    \frac{dy}{dt} &= x(\rho-z)-y,\\
    \frac{dz}{dt} &= xy-\beta z.
\end{align}
Regime 0 uses the classic chaotic parameters
\begin{equation}
    (\sigma,\rho,\beta)=(10,28,8/3),
\end{equation}
whereas regime 1 uses
\begin{equation}
    (\sigma,\rho,\beta)=(10,12,8/3).
\end{equation}
The lower value of $\rho$ changes the bifurcation structure and produces subcritical, fixed-point-like dynamics relative to the chaotic regime. The ground-truth regime switch is therefore the block boundary between $\mu_3$ and $\mu_4$.

Trajectories are integrated with explicit Euler steps using $\Delta t=0.01$. For each regime, the code integrates a batch of independent trajectories, where each trajectory is initialized separately and used to sample repeated trajectory-window observations. In regime 0, initial conditions are sampled from a broad zero-centered Gaussian,
\begin{equation}
    \mathbf{x}(0)\sim\mathcal{N}(\mathbf{0},5^2\mathbf{I}),
\end{equation}
followed by 500 burn-in Euler steps before recording. In regime 1, initial conditions are deliberately biased away from the origin: a base point $(20,22,30)$ is randomly sign-flipped per coordinate and perturbed by Gaussian noise,
\begin{equation}
    \mathbf{x}(0)=\mathbf{s}\odot(20,22,30)+\boldsymbol{\epsilon},
    \qquad
    \boldsymbol{\epsilon}\sim\mathcal{N}(\mathbf{0},6^2\mathbf{I}),
\end{equation}
where $\mathbf{s}\in\{-1,1\}^3$ is sampled independently per trajectory. No additional burn-in is applied in regime 1, so the resulting samples include longer transients under the subcritical dynamics. This initialization scheme makes the two regimes differ not only in parameters but also in the transient portions of state space from which marginal observations are sampled.

Within each regime, long rollouts are used to construct four local marginals. Each marginal sample is a window of length $L=20$ recorded time steps, corresponding to 0.2 simulated time units. The nominal start index for local marginal $\ell\in\{0,1,2,3\}$ within a regime is
\begin{equation}
    a_\ell = 100 + \ell\,(L+\Delta_{\mathrm{gap}}),
    \qquad
    L=20,\quad
    \Delta_{\mathrm{gap}}=150.
\end{equation}
A trajectory-specific temporal jitter $\xi\sim\mathrm{Uniform}\{-15,\ldots,15\}$ is added to these start indices, so the actual window start is $\max(0,a_\ell+\xi)$. The same jitter is shared across the four local marginals of a regime for a given trajectory batch, but differs across independent trajectory initializations. The temporal gaps ensure that marginals are sampled from separated portions of the rollout rather than from nearly overlapping windows.

The global marginal index is
\begin{equation}
    k = r\cdot T_{\mathrm{regime}} + \ell,
    \qquad
    T_{\mathrm{regime}}=4,
\end{equation}
where $r\in\{0,1\}$ is the regime index and $\ell\in\{0,1,2,3\}$ is the local marginal index within that regime. Therefore, marginals $\mu_0,\ldots,\mu_3$ correspond to $\rho=28$, and marginals $\mu_4,\ldots,\mu_7$ correspond to $\rho=12$.

Each sampled window has shape $L\times3$ before storage and is stored as a $3\times L$ trajectory segment for the dataloader. For the flow-matching model, each segment is flattened into a vector in $\R^{3L}$, giving a 60-dimensional observation. The dataset therefore consists of unpaired empirical marginals
\begin{equation}
    \mu_0,\mu_1,\ldots,\mu_7,
\end{equation}
where each marginal contains many independent short trajectory windows from different initial conditions, repeated trial realizations, and trajectory-specific temporal offsets. Consecutive marginals are paired either by index-aligned sampling or, when enabled, by optimal transport between flattened $3\times20$ windows. Ground-truth regime labels are stored from the marginal-to-regime mapping and are used only for evaluation.

\section{NeuralTable Preprocessing Details}
\label{app:neural_preprocessing}

The NeuralTable benchmark uses longitudinal widefield calcium imaging from mice performing a Go/No-Go visual associative learning task. Fluorescence signals are registered to the Allen CCFv3 atlas. Signals are first extracted from 82 bilateral atlas parcels and then averaged into 41 cortical areas. For each trial, we retain frames 11--21 after stimulus onset, corresponding to approximately 1.1--2.1\,s post-stimulus, and flatten the resulting $11\times41$ matrix into a 451-dimensional vector. Activity is z-scored relative to a 1\,s pre-stimulus baseline.

Learning phases are defined from behavior. For each mouse, two change points are identified in the cumulative lick-index curve, defined as the running difference between CS$+$ and CS$-$ anticipatory licking. The first change point marks the transition from early to intermediate learning, when anticipatory licking emerges to both stimuli. The second marks the transition from intermediate to late learning, when differential licking indicates stimulus-specific discrimination.

To align animals with different learning speeds, sessions are aligned within each phase by ordinal position. The normalized interval $[0,1]$ is divided into three equal thirds corresponding to early, intermediate, and late learning. Within each third, session ordinals are placed using \texttt{linspace} according to the maximum number of sessions observed in that phase. This yields $T=22$ day-level marginals: 6 early, 7 intermediate, and 9 late. Each marginal pools trials from all mice at the corresponding aligned ordinal position. Adjacent-session pairs within each mouse form the longitudinal training chain. A 20\% trial-level validation split is used per day-pair so that all sessions are represented in validation.

Behavioral change points are used only to align sessions onto a common longitudinal axis and to define evaluation labels. They are not provided to the velocity model, router $g$, or routing losses during training.

\section{Embryoid Body Preprocessing Details}
\label{app:eb_preprocessing}

The embryoid body benchmark uses single-cell RNA-seq data from embryoid body differentiation. Cells are represented in PCA space, and up to $N=1{,}000$ cells are subsampled per marginal. The dataset contains $T=5$ timepoints. Adjacent timepoints are paired by random permutation during training. Ground-truth stage labels are defined as Pluripotent for timepoints 0--1, Commitment for timepoints 2--3, and Differentiated for timepoint 4. These labels are used only for regime-discovery evaluation.

\section{Evaluation Metrics}
\label{app:evaluation_metrics}

Transport quality is evaluated using sliced Wasserstein distance. For adjacent-step transport,
\begin{equation}
    \mathrm{WD}_1
    =
    \frac{1}{T-1}
    \sum_{k=0}^{T-2}
    W_1
    \left(
    (\phi_{t_k\rightarrow t_{k+1}})_{\#}\mu_k,
    \mu_{k+1}
    \right),
\end{equation}
where $W_1$ is estimated by random one-dimensional projections and $t_k=k/(T-1)$.

Two-hop consistency is measured as
\begin{equation}
    \mathrm{WD}_2
    =
    \frac{1}{|\mathcal{P}_2|}
    \sum_{(i,j)\in\mathcal{P}_2}
    W_1
    \left(
    (\phi_{t_i\rightarrow t_j})_{\#}\mu_i,
    \mu_j
    \right),
    \qquad
    \mathcal{P}_2=\{(i,i+2):0\leq i\leq T-3\}.
\end{equation}
Full-chain error is
\begin{equation}
    \mathrm{WD}_{\mathrm{fc}}
    =
    W_1
    \left(
    (\phi_{t_0\rightarrow t_{T-1}})_{\#}\mu_0,
    \mu_{T-1}
    \right).
\end{equation}

For regime discovery, gating entropy is
\begin{equation}
    H
    =
    -\frac{1}{B}
    \sum_{i=1}^{B}
    \sum_{j=1}^{K}
    w_{ij}\log(w_{ij}+\epsilon).
\end{equation}
Switch rate is the fraction of adjacent temporal segments whose majority-assigned expert changes. Segment-level ARI and NMI are computed by assigning each temporal segment a single predicted label from the majority expert assignment produced by router $g$ and comparing the predicted segment sequence against ground-truth regime labels. These metrics evaluate whether the model recovers temporal boundaries rather than per-sample class labels.

\section{Tables for figures}

\begin{table}[!htbp]
\caption{Lorenz attractor benchmark ($M{=}24$, $T{=}8$, $R{=}2$ regime blocks). Sliced Wasserstein distance ($\downarrow$).}
\label{tab:lorenz}
\centering
\footnotesize
\setlength{\tabcolsep}{3pt}
\resizebox{\columnwidth}{!}{%
\begin{tabular}{llccc}
\toprule
Method & Config.\ & 1-hop WD $\downarrow$ & 2-hop WD $\downarrow$ & Full-chain WD $\downarrow$ \\
\midrule
Gaussian & -- & \textbf{0.030} & \textbf{0.031} & -- \\
Static OT \citep{cuturi2013sinkhorn} & Sinkhorn & 0.035 & 0.033 & -- \\
Indep.\ CFM \citep{tong2024improving} & Per-pair & 49.946 & 58.266 & -- \\
IMMFM \citep{islam2025longitudinal} & Multi-marg.\ & 1.020 & 1.050 & 1.020 \\
FLUX wo Manifold Learning  & Gumbel $K{=}2$ & 1.377 & 2.090 & 1.377 \\
\midrule
\multicolumn{5}{l}{\textit{FLUX (ours --- MoE regime decomposition)}} \\
FLUX & RBF-MLP + Gumbel $K{=}2$ & 0.049 & 0.091 & \textbf{0.179} \\
\bottomrule
\end{tabular}%
}
\end{table}

\begin{table}[t]
\centering
\caption{Embryoid body differentiation ($T=5$ timepoints, 3 stages). Sliced Wasserstein distance ($\downarrow$). Regime metrics in Table~\ref{tab:regime}.}
\label{tab:eb}
\resizebox{\textwidth}{!}{
\begin{tabular}{llccc}
\toprule
\textbf{Method} & \textbf{Config.} & \textbf{1-hop WD $\downarrow$} & \textbf{2-hop WD $\downarrow$} & \textbf{Full-chain WD $\downarrow$} \\
\midrule
Gaussian & -- & \textbf{0.061} & \textbf{0.064} & -- \\
Static OT \citep{cuturi2013sinkhorn} & Sinkhorn & 0.791 & 0.834 & -- \\
Vanilla CFM \citep{lipman2023flow} & Eucl., shared & 0.267 & 0.309 & -- \\
Indep. CFM \citep{tong2024improving} & Per-pair & 0.364 & 0.458 & -- \\
IMMFM \citep{islam2025longitudinal} & Multi-marg. & 0.128 & 0.181 & 0.368 \\
FLUX wo Manifold Learning  & Gumbel $K=3$ & 0.126 & 0.173 & \textbf{0.126} \\
\midrule
\multicolumn{5}{l}{\textit{FLUX (ours --- MoE regime decomposition)}} \\
FLUX & RBF-MLP + Gumbel $K=3$ & 0.210 & 0.263 & 0.450 \\
\bottomrule
\end{tabular}
}
\end{table}

\begin{table}[t]
\centering
\caption{NeuralTable benchmark (widefield Ca$^{2+}$ imaging, 12 mice, 451-d, 22 marginals). Sliced Wasserstein distance ($\downarrow$).}
\label{tab:neural_table}
\resizebox{\textwidth}{!}{
\begin{tabular}{llccc}
\toprule
\textbf{Method} & \textbf{Config.} & \textbf{1-hop WD $\downarrow$} & \textbf{2-hop WD $\downarrow$} & \textbf{Full-chain WD $\downarrow$} \\
\midrule
Gaussian & -- & \textbf{0.001} & \textbf{0.001} & -- \\
Static OT \citep{cuturi2013sinkhorn} & Sinkhorn & 0.007 & 0.007 & -- \\
Vanilla CFM \citep{lipman2023flow} & Eucl., shared & 0.002 & 0.002 & -- \\
Indep. CFM \citep{tong2024improving} & Per-pair & 0.006 & 0.007 & -- \\
IMMFM \citep{islam2025longitudinal} & Multi-marg. & \textbf{0.001} & 0.002 & \textbf{0.006} \\
FLUX wo Manifold Learning  & Gumbel $K=3$ & 0.147 & 0.187 & 0.147 \\
\midrule
\multicolumn{5}{l}{\textit{FLUX (ours --- MoE regime decomposition)}} \\
FLUX & RBF-MLP + Gumbel $K=3$ & 0.002 & 0.004 & 0.030 \\
\bottomrule
\end{tabular}
}
\end{table}

\begin{table}[t]
\centering
\caption{Segment-level regime discovery (ARI $\uparrow$, NMI $\uparrow$) across all benchmarks with ground-truth labels.}
\label{tab:regime}

\begin{tabular}{lcccccc}
\toprule
& \multicolumn{2}{c}{\textbf{Lorenz}} & \multicolumn{2}{c}{\textbf{NeuralTable}} & \multicolumn{2}{c}{\textbf{Embryoid Body}} \\
\cmidrule(lr){2-3}\cmidrule(lr){4-5}\cmidrule(lr){6-7}
\textbf{Method} & \textbf{ARI} & \textbf{NMI} & \textbf{ARI} & \textbf{NMI} & \textbf{ARI} & \textbf{NMI} \\
\midrule
K-Means & 0.054 & 0.316 & 0.487 & 0.702 & 0.000 & 0.000 \\
GMM & -0.077 & 0.163 & 0.062 & 0.189 & 0.571 & 0.800 \\
PCA + K-Means & 0.054 & 0.316 & 0.487 & 0.702 & -0.333 & 0.269 \\
Spectral & 0.432 & 0.508 & 0.000 & 0.000 & 0.000 & 0.000 \\
rSLDS & 1.000 & 1.000 & 0.000 & 0.000 & -- & -- \\
SRNN & 1.000 & 1.000 & 0.000 & 0.000 & -- & -- \\
\midrule
FLUX without Geometric Learning & 0.000 & 0.000 & 0.200 & 0.235 & 0.571 & 0.800 \\
\midrule
\multicolumn{7}{l}{\textit{FLUX (ours) --- MoE Gating}} \\
FLUX Gumbel & 1.000 & 1.000 & 0.461 & 0.589 & 0.571 & 0.800 \\
\bottomrule
\end{tabular}
\end{table}
\section{Baselines}
\label{app:baseline_notes}

\paragraph{Gaussian baseline.}
For each marginal, a Gaussian distribution is fitted using the empirical mean and covariance of the observed samples. Samples are then drawn from the fitted Gaussian for evaluation.

\paragraph{Linear interpolation.}
Random endpoint pairs are sampled between adjacent marginals, and transport is defined by linear interpolation between paired samples. This baseline measures how much of the adjacent marginal transition can be explained without a learned velocity field. In datasets where the evaluation samples lie exactly on the interpolated endpoints, this baseline can produce near-zero adjacent-step WD, but it does not define a shared longitudinal ODE or a learned full-chain transport model.

\paragraph{Static OT.}
Static optimal transport is computed independently for each adjacent pair using Sinkhorn-regularized optimal transport. Because this method does not define a shared continuous-time velocity field, full-chain ODE-based evaluation is not reported.

\paragraph{Vanilla CFM.}
Vanilla CFM trains a single shared velocity network using Euclidean interpolation paths across all adjacent marginal pairs.

\paragraph{Independent CFM.}
Independent CFM trains a separate conditional flow-matching model for each adjacent marginal pair. It provides a strong short-horizon reference but does not share parameters across the full chain.

\paragraph{IMMFM and FLUX without Geometric Learning.}
IMMFM provides a longitudinal Euclidean multi-marginal flow-matching baseline. FLUX without Geometric Learning augments the same adjacent-marginal transport setting with the FLUX MoE velocity architecture, router $g$, Gumbel-Softmax expert selection, and routing losses, but removes metric learning and the bend network. Its conditional paths and velocity targets are Euclidean. This ablation isolates the contribution of geometry-aware conditional paths while keeping the routing objective fixed.
\paragraph{Regime-discovery baselines.}
K-Means, GMM, PCA + K-Means, and Spectral Clustering are applied to the observed feature representation or to PCA-reduced features. rSLDS and SRNN serve as switching-dynamics baselines. Because these methods require sequential inputs, they are applied to available ordered trajectories when present and to pseudo-sequences constructed from adjacent-marginal pairings otherwise.

\section{Reproducibility Checklist}
\label{app:reproducibility_checklist}

All experiments use the same modular pipeline:
\begin{enumerate}
    \item construct or load longitudinal marginals;
    \item train the geometry backend;
    \item train the bend network;
    \item train the single-velocity or MoE velocity model;
    \item evaluate one-hop, two-hop, and full-chain transport;
    \item evaluate segment-level regime discovery where labels are available.
\end{enumerate}
Synthetic benchmarks include fixed random seeds and full data-generation scripts. Biological benchmarks include preprocessing scripts, train/validation splits, model checkpoints, and table-generation scripts. Code will be released upon publication.

\section{Stanford Bunny Mesh Dataset and Dimensionality Ablation}
\label{app:stanford_bunny}

\subsection{Data Source and Surface Geometry}
\label{app:bunny_geometry}

The Stanford Bunny control uses a simplified triangular mesh, \texttt{bunny\_simp.obj}, containing approximately 2,500 vertices and associated triangle faces. The mesh is loaded with the Python bindings for \texttt{libigl}, which provides access to mesh operations and surface-distance queries. This dataset is used as a controlled manifold benchmark because the target state space is known: all target samples lie exactly on a two-dimensional triangulated surface embedded in three dimensions before the high-dimensional projection step.

\subsection{Geodesic Marginal Construction}
\label{app:bunny_marginals}

We construct $K=8$ ordered empirical marginals $\mu_0,\ldots,\mu_7$ directly on the mesh surface. In the geodesic mode used for the dimensionality ablation, heat-geodesic distances are first computed from the mesh centroid. The start vertex is selected as the vertex farthest from the centroid, and the end vertex is selected as the vertex farthest from the start under the same geodesic-distance operator. For the simplified bunny mesh, this procedure gives a geodesic path of approximately 818 mesh units, from vertex 851 to vertex 2269.

Eight waypoint positions are placed at evenly spaced geodesic progress values
\begin{equation}
    s_k \in \{0.00,0.14,0.29,0.43,0.57,0.71,0.86,1.00\},
    \qquad k=0,\ldots,7.
\end{equation}
For each waypoint, mesh faces are sampled with probability proportional to face area multiplied by a Gaussian falloff in geodesic distance from the waypoint,
\begin{equation}
    p(f\mid k) \propto \mathrm{area}(f)
    \exp\left[-\frac{1}{2}\left(\frac{d_{\mathcal{M}}(f,s_k)}{b}\right)^2\right],
    \qquad b=30,
\end{equation}
where $d_{\mathcal{M}}(f,s_k)$ denotes the geodesic distance from face $f$ to waypoint $s_k$. After a face is selected, a random barycentric coordinate is drawn uniformly on the triangle simplex. This produces samples that lie exactly on the triangular mesh. Each marginal contains 4,000 sampled surface points.

\subsection{High-Dimensional Orthonormal Embedding}
\label{app:bunny_highd}

To test sensitivity to ambient dimension while keeping the intrinsic manifold fixed, the 3D mesh samples are embedded into $\mathbb{R}^{D}$ for
\begin{equation}
    D\in\{3,5,10,20,50,100\}.
\end{equation}
For each dimension, a Gaussian random matrix $G\in\mathbb{R}^{D\times3}$ is generated with a fixed seed. QR decomposition gives an orthonormal basis $Q$, and the embedding matrix is defined as
\begin{equation}
    A = Q_{:,1:3}^{\top}\in\mathbb{R}^{3\times D}.
\end{equation}
A 3D surface point $\mathbf{x}\in\mathbb{R}^{3}$ is mapped to
\begin{equation}
    \mathbf{x}^{(D)} = \mathbf{x}A \in \mathbb{R}^{D}.
\end{equation}
Because the rows of $A$ are orthonormal, Euclidean distances among the embedded 3D points are preserved. The intrinsic surface remains three-dimensional, but its co-dimension increases from $D-3$. This isolates a common biological regime: observations may be high-dimensional even when the population states occupy a much lower-dimensional curved manifold.

\subsection{Held-Out Marginal Protocol}
\label{app:bunny_holdout}

Two of the eight marginals are held out from velocity training: $\mu_1$ and $\mu_6$, corresponding to geodesic progress values 0.14 and 0.86. The velocity model is trained only on the retained marginals
\begin{equation}
    \{\mu_0,\mu_2,\mu_3,\mu_4,\mu_5,\mu_7\}.
\end{equation}
Evaluation still integrates the learned dynamics through all eight target times, including the two held-out marginals. This protocol tests whether the model can interpolate to unseen surface-localized distributions, rather than only matching the observed training marginals.

\subsection{Models Compared}
\label{app:bunny_models}

\paragraph{IMMFM baseline.}
The Euclidean baseline is an independent marginal-matching flow-matching model with a single MLP velocity network. The model uses hidden dimension 512, depth 6, and approximately 2.5M parameters. It is trained on the six retained marginals using piecewise conditional flow matching across adjacent retained marginals. Training uses a composition loss with weight 0.1, a $\sigma_M=2.0$ schedule, batch size 128, learning rate $10^{-4}$, and at most 600 epochs with early-stopping patience 60. This model has no access to the mesh, the learned metric, or the off-manifold penalty.

\paragraph{FLUX with RBF/RBF-MLP geometry.}
The geometry-aware pipeline uses three stages. Stage 0 trains the geometry backend on all eight marginal point clouds. For $D\leq5$, the backend is the direct RBF metric with 100 k-means centers and bandwidth $10\sqrt{D/3}$. For $D>5$, the backend is an RBF-MLP deep-kernel metric with a four-dimensional latent feature map and a 3-layer MLP backbone of width 256. The geometry stage is trained for up to 200 epochs with early-stopping patience 30, batch size 256, and learning rate $10^{-2}$. Stage 1 trains the bend network on the six retained marginals using the frozen metric. The bend network is a 3-layer MLP with width 512 and approximately 557K parameters, trained for up to 400 epochs with early-stopping patience 50, batch size 128, and learning rate $10^{-4}$. Stage 2 trains the factorized velocity model on the six retained marginals. The velocity model has $K=2$ experts, Gumbel-Softmax routing with temperature annealed from 1.0 to 0.1 at rate 0.98, three hidden layers of width 512, and a 64-dimensional time embedding. It uses OT coupling, quadratic interpolation, previous-step conditioning, load-balancing regularization, z-loss, confidence penalty, clustering penalty, temporal total-variation penalty, and composition loss. The velocity stage is trained for up to 800 epochs with early-stopping patience 80, cosine learning-rate scheduling, batch size 128, and learning rate $10^{-4}$.

\subsection{Sequential Integration and Metrics}
\label{app:bunny_eval}

All models are evaluated using the same sequential integration protocol. Starting from 2,000 samples from $\mu_0$, the learned velocity field is integrated across each consecutive interval $(t_k,t_{k+1})$ using Euler integration with 100 steps per segment. Generated samples are recorded at all eight target times. For factorized models, the segment index is provided to the router when segment conditioning is enabled.

Generated $D$-dimensional samples are projected back to 3D using the pseudoinverse of the embedding matrix,
\begin{equation}
    \widehat{\mathbf{x}}^{(3)} = \widehat{\mathbf{x}}^{(D)} A^{\dagger}.
\end{equation}
The first transport metric is coordinate-wise Wasserstein distance,
\begin{equation}
    \mathrm{WD}_{\mathrm{3D}}(k)
    =
    \frac{1}{3}\sum_{r=1}^{3}
    W_1\left(\widehat{X}^{(3)}_{k,r},X^{(3)}_{k,r}\right),
\end{equation}
where $\widehat{X}^{(3)}_{k,r}$ and $X^{(3)}_{k,r}$ denote the generated and target coordinates for marginal $k$. We report held-out WD averaged over $k\in\{1,6\}$, training-marginal WD averaged over $k\in\{2,3,4,5,7\}$, and all-marginal WD averaged over $k=0,\ldots,7$.

The second metric is surface deviation. For each projected 3D generated point, the signed distance to the closest mesh surface point is computed with \texttt{igl.signed\_distance}. The reported deviation is
\begin{equation}
    \mathrm{SurfDev}
    =
    \frac{1}{K}\sum_{k=0}^{K-1}
    \mathbb{E}_{\widehat{\mathbf{x}}\sim \widehat{\mu}_k}
    \left[\left|d_{\mathrm{mesh}}(\widehat{\mathbf{x}})\right|\right].
\end{equation}
This metric directly tests whether the learned transport remains on the known surface. A model that follows the mesh should have low surface deviation, whereas a Euclidean model that cuts through the bunny interior or leaves the surface should have larger deviation.

\end{document}